# Using DUCK-Net for Polyp Image Segmentation

Razvan-Gabriel Dumitru[1], Darius Peteleaza[2], Catalin Craciun[3]

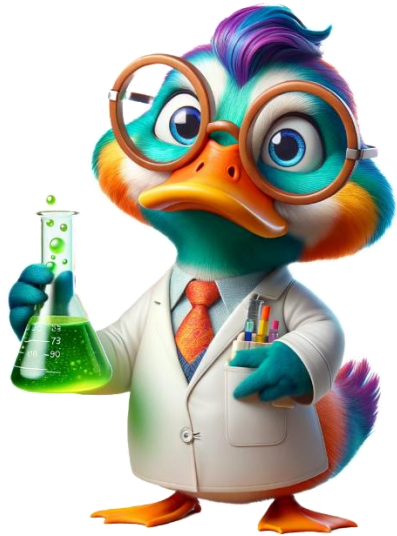

## Abstract

This paper presents a novel supervised convolutional neural network architecture, "DUCK-Net", capable of effectively learning and generalizing from small amounts of medical images to perform accurate segmentation tasks. Our model utilizes an encoder-decoder structure with a residual downsampling mechanism and a custom convolutional block to capture and process image information at multiple resolutions in the encoder segment. We employ data augmentation techniques to enrich the training set, thus increasing our model's performance. While our architecture is versatile and applicable to various segmentation tasks, in this study, we demonstrate its capabilities specifically for polyp segmentation in colonoscopy images. We evaluate the performance of our method on several popular benchmark datasets for polyp segmentation, Kvasir-SEG, CVC-ClinicDB, CVC-ColonDB, and ETIS-LARIBPOLYPDB showing that it achieves state-of-the-art results in terms of mean Dice coefficient, Jaccard index, Precision, Recall, and Accuracy. Our approach demonstrates strong generalization capabilities, achieving excellent performance even with limited training data. The code is publicly available on GitHub: https://github.com/RazvanDu/DUCK-Net

[1]Independent Researcher, Sibiu, 550201, Romania, razvandumm@gmail.com
[2]Independent Researcher, Sibiu, 550201, Romania, peteleaza.darius@gmail.com
[3]Independent Researcher, Medias, 551112, Romania, craciun.catalin.c@gmail.com

# Introduction

Colorectal cancer (CRC) is a leading cause of cancer mortality globally [1]. Most colorectal cancers evolve from adenomatous polyps, making early detection and removal of polyps critical for CRC prevention and treatment [2]. Colonoscopy is the gold standard for detecting and removing polyps before they develop into CRC [3]. However, accurately identifying and segmenting polyps during colonoscopy is a complex task due to the diversity of polyps in terms of shape, size, and texture. This can lead to missed or misdiagnosed polyps, which can seriously harm patient health.

Machine learning (ML) algorithms, particularly convolutional neural networks (CNNs), have shown promising results in medical image segmentation and have been applied to polyp detection and segmentation [4, 5]. While deep learning (DL) algorithms can achieve high precision, they typically require large amounts of labeled data [6, 38, 39], which can be costly and time-consuming to obtain [40].

In an effort to improve the accuracy and efficiency of polyp segmentation, researchers have developed various deep learning (DL) architectures that employ different techniques to address this complex task. Examples of DL architectures used for polyp segmentation include U-Net [7], FCN [8], and their variants, such as U-Net++ [9], Modified U-Net (mU-Net) [41], ResUNet++ [10], and H-DenseUNet [11]. While these methods can achieve precise segmentation results, their performance may be less robust when faced with a wide range of polyp characteristics.

In this study, we present a novel supervised convolutional neural network architecture for image segmentation that uses the encoder-decoder structure of the U-Net [7] architecture with some significant differences. The key feature of our architecture is the combination between our custom-designed convolutional block and the residual downsampling. The convolutional block enables our model to accurately locate and predict the borders of polyps with a small margin of error. By incorporating residual downsampling, the model can utilize initial image information at each resolution level in the encoder segment, further improving its performance. Also, we have used DeepLabV3 atrous convolutions [12] for capturing spacial information and the residual block of ResUNet++ [10] for enhanced feature extraction.

The main contributions of this paper are:

- Our custom-built convolutional block, DUCK (Deep Understanding Convolutional Kernel), allows more in-depth feature selection, enabling the model to locate the polyp target accurately and correctly predict its borders.
- Our method uses residual downsampling, which allows it to use the initial image information at each resolution level in the encoder segment. This way, the network always has the original field of view alongside the processed input image.
- Our model does not use external modules and was only trained on the target dataset (no pre-training of any kind)
- Our method accurately identifies polyps regardless of number, shape, size, and texture.
- Extensive experiments prove that our method achieves good performance and leads existing methods on several benchmark datasets.

## Related work

### Convolutional Neural Networks

Automatic polyp segmentation is crucial in clinical practice to reduce cancer mortality rates. Medical image segmentation tasks usually employ convolutional neural networks, and several widely utilized architectures have been applied to this problem.

One such architecture is U-Net [7], an encoder-decoder model developed initially for biomedical image segmentation. U-Net exhibits the advantage of being relatively simple and efficient while still achieving good performance on various medical image segmentation tasks. However, it may struggle with more complex or varied input images, and alternative methods may be more suitable in these cases.

PraNet [32] is a CNN architecture specifically designed for automatic polyp segmentation in colonoscopy images. It employs a parallel partial decoder to extract high-level features from the images and generate a global map as initial guidance for the following processing steps. Furthermore, it utilizes a reverse attention module to mine boundary cues, which helps to establish the relationship between different regions of the images and their boundaries. PraNet also incorporates a recurrent cooperation mechanism to correct misaligned predictions and improve segmentation accuracy. The results of the evaluations indicate that PraNet significantly improves the segmentation accuracy and has an advantage in terms of real-time processing efficiency, reaching a speed of about 50 frames per second.

DeepLabV3+ [13] is an extension of the DeepLabV3 [12] architecture for semantic image segmentation. It employs atrous convolutions, which allow for a dilated field of view and the extraction of features at multiple scales to improve the capture of long-range contextual dependencies. This approach enables the more accurate segmentation of objects with complex shapes or large-scale variations but also requires more computation and may be slower to train and infer.

HRNetV2 [14, 15] is a CNN architecture for human pose estimation that utilizes a fully connected style-like architecture to share multi-scale information between layers at different resolutions. This architecture can improve performance on small or blurry objects but may be more prone to overfitting and require more data to achieve good performance.

Other CNNs designed explicitly for automatic polyp segmentation include ResUNet [16], which incorporates residual blocks to enhance location information for polyps, and HarDNet-DFUS [17], which combines a custom-built encoder block called HarDBlock with the decoder of Lawin Transformer to improve accuracy and inference speed. ResUNet can leverage the powerful expressive capacity of residual blocks but may require more data and computation to achieve good performance. HarDNet-DFUS is designed with real-time prediction in mind but may sacrifice some accuracy in favor of faster inference.

ColonFormer [18] utilizes attention mechanisms in the encoder and includes a refinement module with attention on the x and y axis at different resolutions to achieve a more refined output while maintaining a decoder similar to the classical U-Net. Attention mechanisms can be effective for handling large or complex input images but may require more computation and be more challenging to optimize than other methods.

MSRF-Net [21] is a CNN architecture specifically designed for medical image segmentation. It utilizes a unique Dual-Scale Dense Fusion (DSDF) block to exchange multi-scale features with varying receptive fields, allowing the preservation of resolution and improved information flow. The MSRF sub-network then employs a series of these DSDF blocks to perform multi-scale fusion, enabling the propagation of high-level and low-level features for accurate segmentation. However, one limitation of this method is that it may not perform well on low-contrast images.

### Transformers

While the previously mentioned methods have achieved good results for automatic polyp segmentation, other approaches that utilize transformers in the encoder perform particularly well on this task. These models typically use a pre-trained vision transformer as an encoder trained on a large dataset, such as ImageNet [22], to extract relevant features from the input image. These features are then fed to the decoder, which processes multi-scale features and combines them into a single, final output. Examples of such approaches include FCN-Transformer [19] and SSFormer-L [20], which have achieved state-of-the-art (SOTA) performance on the Kvasir Segmentation Dataset at the time of their release.

The use of Transformers has gained traction in the field of Computer Vision (CV) in the past years, as they have been widely used in the field of Natural Language Processing (NLP) and have shown spectacular results in retaining the global context of the subject at hand. Vision-Transformers (ViT) [34], like their NLP counterparts, make use of a mechanism called Attention [33], which aggregates global context to extract relevant information from large image patches.

While ViTs [34] seem to perform well in the CV field, traditional CNN methods, like the EfficientNetV2 [35], outperformed them in popular image classifications datasets, such as ImageNet [22] or CIFAR-10 [36], proving that more efficient CNN methods can still be developed.

As such, our proposed method explores the benefits of traditional CNNs over ViT-based architectures in biomedical image segmentation and how they can still yield substantial improvements in the accuracy metrics.

Overall, this field is an active area of research, with various approaches being proposed and evaluated. Thus, further research is needed to determine the models' optimal design and training strategies. It is essential to carefully consider the trade-offs between accuracy, computational efficiency, and other performance metrics when selecting a method for a specific application.

# Methodology

The proposed polyp segmentation solution consists of two novel main components. The first is a novel convolutional block called DUCK that uses six variations of convolutional blocks in parallel to allow the network to train whichever it deems best. While the novel convolutional block allows the network to train the most critical parts precisely, one drawback is that it crushes fine details for the subsequent layers. The second novel contribution keeps the low-level details by adding a secondary U-Net [7] downscaling layer that does not process the image, so it keeps the low-level details intact. We will present each in detail, explaining the high-level architecture and the convolutional blocks.

## Model architecture

Our proposed architecture (Figure 1) uses the encoder-decoder formula of the U-net [7] architecture with three significant differences.

Firstly, we replace the pair of 3x3 convolutional blocks classically used by U-Net [7] with our novel DUCK block at each step except the last one. This allows the model to capture more details at each step while sacrificing finer low-level details. The exact details of the block and the explanation behind how it works are detailed below. For the last downsampling part, we chose to go with four Residual Blocks [10] because the image size after being downscaled five times is $11$ ($\frac{352}{2^5}$), which is smaller than the largest simulated kernel size of the DUCK. Thus, it would not be able to take full advantage at such a small scale.

Secondly, to address the issues caused by the novel block, such as losing fine details, we have implemented a secondary downscaling layer that does not implement any convolutional processing. The output from each step of this layer is then fed into the main downscaling layer using addition. We employ 2D 2x2 convolutions with strides of 2 to downscale the image. This behaves better than max pooling as the model can learn the essential parts to keep.

Lastly, we used addition instead of concatenation every time we combined two outputs, similar to LinkNet [37], as we observed that it produces better results using less memory and computational resources. This also means that at each step, we need to have half of the number of parameters on the upscaling part to match the output size of the downscaling part.

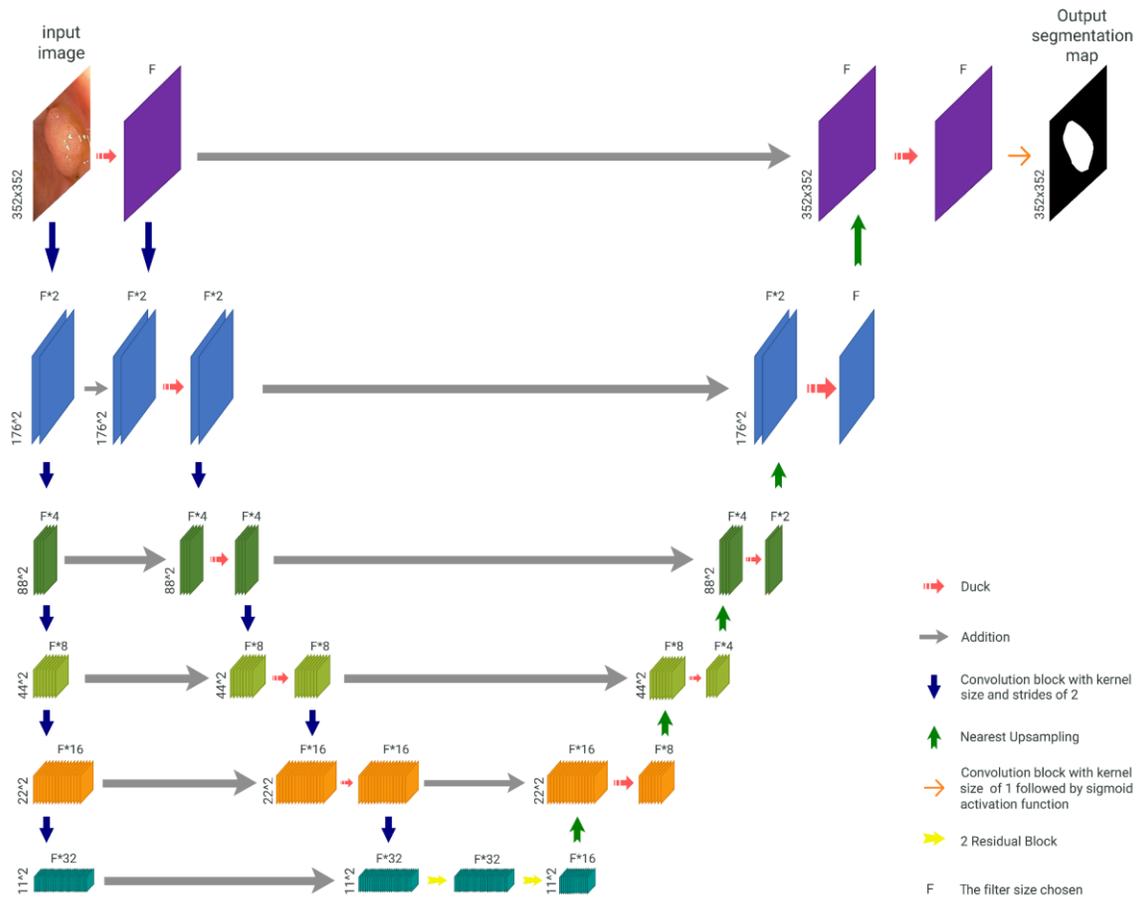

**Figure 1.** DUCK-Net architecture

In our study, we utilized a parameter, F (filter size), to modify the depth of convolutional layers. Through comprehensive experimentation, we determined that a model incorporating 17 filters serves as an optimal representation of a smaller model, while a model incorporating 34 filters represents a larger model effectively.

**Block components**

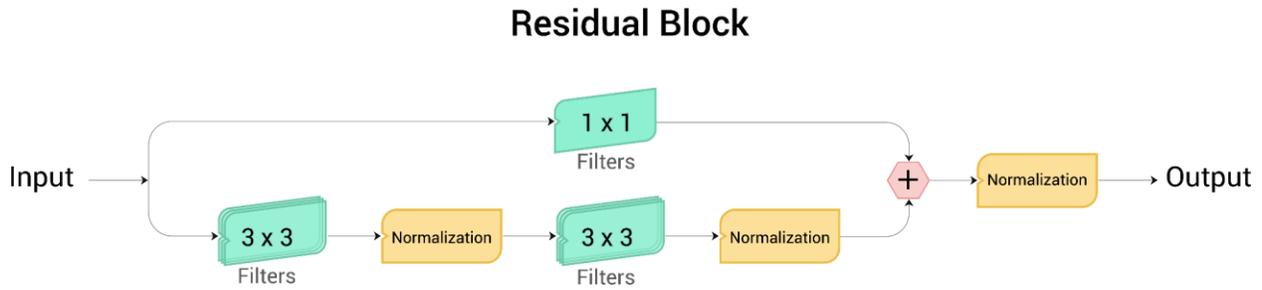

**Figure 2.** Residual block

The Residual block (Figure 2), first introduced in ResUNet++ paper [10], is the first component in our novel DUCK. Its purpose is to understand the small details that make a polyp. While using multiple small convolutions is usually a good idea, having too many can mean that the network has difficulty training and understanding what features to look for. We use combinations of one, two, and three Residual blocks to simulate kernel sizes of 5x5, 9x9, and 13x13.

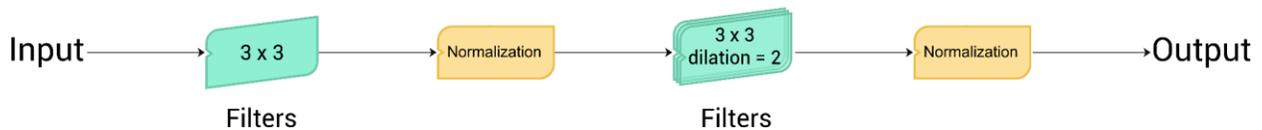

**Figure 3.** Midscope block

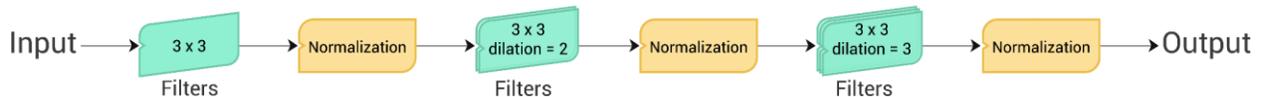

**Figure 4.** Widescope block

Our novel Midscope (Figure 3) and Widescope (Figure 4) blocks use dilated convolutions to reduce the parameters needed to simulate larger kernels while allowing the network to understand higher-level features better. They work by spreading the nine cells that would typically be in a 3x3 kernel over a larger area. These two blocks aim to learn prominent features that only require a little attention to detail, as the dilation effect has the side effect of losing information. The Midscope cell simulates a kernel size of 7x7, and the Widescope simulates a kernel size of 15x15.

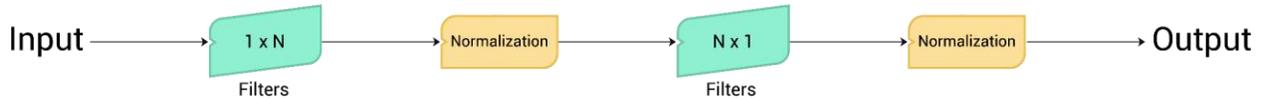

**Figure 5.** Separated block

The Separated block (Figure 5) is our third way of simulating big kernels. The main idea behind it is that combining a 1xN kernel with an Nx1 kernel results in a behavior similar to an NxN kernel. However, this method encounters a drawback related to the concept known as "diagonality". Essentially, diagonality implies the capacity of a convolutional layer to capture and sustain spatial details linked to diagonal patterns in an image, a feature intrinsic to the structure of a conventional NxN convolutional kernel. It retains these diagonal elements owing to its bidimensional characteristics, enabling it to capture spatial connections in both vertical and horizontal directions, which also encompasses diagonal aspects. Yet, the distinctive processing approach of separable convolutions (1xN followed by Nx1), where filters operate on one dimension at a time, potentially obstructs their capacity to efficiently encode diagonal features. This leads to the so-called "loss of diagonality". Such diagonal relationships can prove useful for detecting specific intricate patterns or shapes within an image, hence the other blocks are designed to compensate.

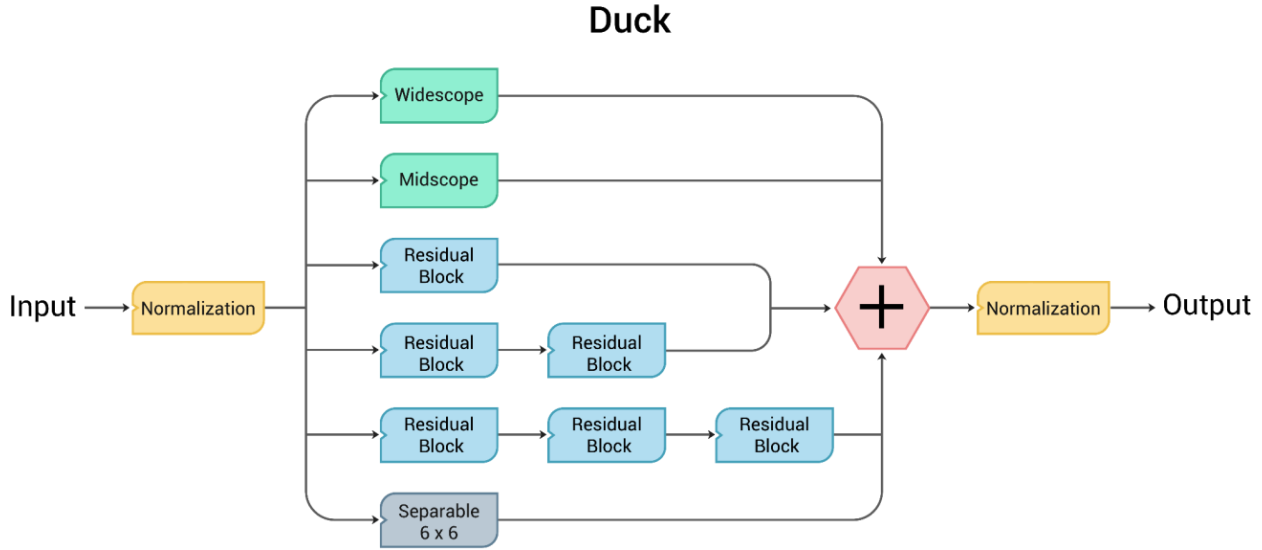

**Figure 6.** DUCK block

DUCK (Figure 6) is our novel convolutional block that combines the previously mentioned blocks, all used in parallel so that the network can use the behavior it deems best at each step. The idea behind it is that it has a wide variety of kernel sizes simulated in three different ways. This means that the network can decide how to compensate for the drawbacks of one way to simulate a kernel over another. Having a variety of kernel sizes means it can find the general area of the target while also finding the edges correctly. We incorporated a one-two-three combination of residual blocks based on empirical observations suggesting no significant performance gains from multiple instances of Midscope, Widescope, and Separable blocks. Essentially, the computational resources required for these additions did not justify the marginal improvements in results. The result is a novel block that searches for low-level and high-level features simultaneously with auspicious results.

### Model evaluation

Accurate evaluation is crucial for determining the effectiveness of various neural network architectures. Several metrics have been proposed for this purpose, and we have chosen to focus on five of the most widely used: the Dice Coefficient, Jaccard Index, Precision, Recall, and Accuracy.

1. The Dice Coefficient, also known as the F1 score, is a measure of the overlap between two sets, with a range of 0 to 1. A value of 1 indicates a perfect overlap, while 0 indicates no overlap.

2. The Jaccard Index, similar to the Dice Coefficient, measures the overlap between two sets but is expressed as a ratio of the size of the intersection to the size of the union of the sets.
3. Precision is a measure of the positive predictive value of a classifier or the proportion of true positive predictions among all positive predictions.
4. Recall, also known as sensitivity or true positive rate, measures the proportion of true positive predictions among all actual positive instances.
5. Accuracy is the overall correct classification rate or the proportion of correct predictions made by the classifier out of all predictions made.

$$Dice\ coefficient = \frac{2TP}{2TP + FP + FN} \tag{1}$$

$$Jaccard\ index = \frac{TP}{TP + FP + FN} \tag{2}$$

$$Precision = \frac{TP}{TP + FP} \tag{3}$$

$$Recall = \frac{TP}{TP + FN} \tag{4}$$

$$Accuracy = \frac{TP + TN}{TP + TN + FP + FN} \tag{5}$$

The Dice loss is a loss function commonly used in medical image segmentation tasks. It uses the Dice Coefficient, which measures the overlap between two sets. In the context of image segmentation, the Dice loss can be used to penalize the model for incorrect or incomplete segmentation of objects in the image.

Using the Dice loss for medical image segmentation has several benefits:

1. The Dice Coefficient is widely used to evaluate the performance of image segmentation models, so using the Dice loss helps optimize the model for this metric.
2. The Dice loss can handle class imbalance, which is often a concern in medical image segmentation, where some classes may be much more prevalent than others.
3. The Dice loss is differentiable, which allows it to be used in conjunction with gradient-based optimization algorithms.

The Dice loss is calculated as follows:

$$Dice\ Loss = 1 - Dice\ Coefficient \qquad (6)$$

# Experiments

## Implementation details

To ensure fairness and reproducibility in our comparisons, we used identical training, validation, and testing sets for all models evaluated in our study. Specifically, each dataset was randomly split into three subsets: training, validation, and testing, with an 80:10:10 percent ratio. The motivation behind choosing a random data split was to ensure that the selection process was unbiased and that the comparison across different models was as fair as possible. We provide the split datasets in the "Data Availability" section so our results are easily reproducible.

We designed our experimental setup to validate our model's state-of-the-art performance on unseen data while showcasing its ability to generalize across different contexts. First, we conducted tests on each dataset independently and compared our model's performance with the other methods. Then, to prove the generalization capabilities of our model, we trained the model on one dataset and tested it on another, namely the Kvasir-SEG [28] and CVC-ClinicDB [29] datasets and vice versa. This way, we could effectively gauge its adaptability and predictive accuracy on novel, unseen data. This cross-dataset testing yielded strong results, emphasizing our model's generalization capabilities, even in the absence of any extra pre-training data.

We trained our model to predict binary segmentation maps for RGB images. To reduce the computational cost, the images are rescaled to 352 x 352 pixels. This convention was set by several published papers [18, 19, 20, 32]. Due to the aliasing issues with rescaling images [23], we used a Lanczos filter [26] to preserve the quality. We used the RMSprop [24] optimizer, with a learning rate of 0.0001. We trained our model with a batch size of 4 for 600 epochs. We used Tensorflow [25] as our framework to implement the architecture and trained the model using an NVIDIA A100 GPU.

## Data augmentation

We implemented data augmentation on the training set, significantly improving the model's generalization capabilities to the point where regularization techniques such as dropout were unnecessary. The library used to implement the augmentations is Albumenations [27]. This involved randomly applying transformations to the training images, resulting in significantly different variations from the original images and helping the model better generalize to unseen data.

Before each epoch, we randomly augmented the training input using augmentations inspired by previous work [19] but modified to fit the specific needs of our model. The augmentation techniques we used are:

1. Horizontal and vertical flips
2. Color jitter with a brightness factor uniformly sampled from [0.6, 1.6], a contrast of 0.2, a saturation factor of 0.1, and a hue factor of 0.01
3. Affine transforms with rotations of an angle sampled uniformly from [−180°, 180°], horizontal and vertical translations each of a magnitude sampled uniformly from [-0.125, 0.125], scaling of a magnitude sampled uniformly from [0.5, 1.5] and shearing of an angle sampled uniformly from [−22.5°, 22°].

Out of these augmentations, the color jitter was applied only to the image, while the rest were applied consistently to both the image and the corresponding segmentation map.

### Datasets

We perform experiments on the most popular four datasets for polyp segmentation: Kvasir-SEG [28], CVC-ClinicDB [29], CVC-ColonDB [30], and ETIS-LARIBPOLYPDB [31].

- **The Kvasir-SEG dataset** [28] contains 1000 polyp images and their corresponding ground truth, with different resolutions ranging from 332 x 487 to 1920 x 1072 pixels;
- **The CVC-ClinicDB dataset** [29] contains 612 polyp images and their corresponding ground truth, with a resolution of 384 x 288 pixels;
- **The ETIS-LARIBPOLYPDB dataset** [31] contains 196 polyp images and their corresponding ground truth, with a resolution of 1255 x 966 pixels;
- **The CVC-ColonDB dataset** [30] contains 380 polyp images and their corresponding ground truth, with a resolution of 574 x 500 pixels;

### Results

The tables below show the comparison of different methods using mean Dice, Jaccard index, Precision, Recall, and Accuracy metrics. We also included the calculation of standard deviation (SD) in our analysis to strengthen our evaluation of model performance. This measure provides insight into the variability of the used metrics among different models, thus giving us an understanding of the potential range of performance when employing these methods. This statistical perspective complements the raw performance figures, offering a more comprehensive view of the model's performance consistency and reliability.

To ensure a fair comparison, we utilized image augmentations for the base U-Net [7] model. These augmentations were consistent with those used in our model. Also, to provide a clearer understanding of the results, we included information in the tables regarding which methods were pre-trained.

| Kvasir-SEG | | | | | |
|---|---|---|---|---|---|
| **Method** | **DSC** | **Jaccard** | **Precision** | **Recall** | **Accuracy** |
| U-Net [7] (with our augmentations) | 0.8655 | 0.7629 | 0.8593 | 0.8718 | 0.9563 |
| HRNetV2 [14, 15] | 0.8530 | 0.7438 | 0.8778 | 0.8297 | 0.9539 |
| PraNet [32] (pre-trained) | 0.9094 | 0.8339 | 0.9599 | 0.8640 | 0.9738 |
| HarDNet-DFUS [17] (pre-trained) | 0.8626 | 0.7584 | 0.9351 | 0.8005 | 0.9583 |
| MSRF-Net [21] | 0.8508 | 0.7404 | 0.8993 | 0.8074 | 0.9543 |
| FCN-Transformer [19] (pre-trained) | 0.9220 | 0.8554 | 0.9238 | 0.9203 | 0.9749 |
| **OURS (no pre-training, 17 filters)** | 0.9343 | 0.8769 | 0.9350 | 0.9337 | 0.9789 |
| **OURS (no pre-training, 34 filters)** | **0.9502** | **0.9051** | **0.9628** | **0.9379** | **0.9842** |
| Standard Deviation (between methods) | 0.0373 | 0.0615 | 0.0349 | 0.0519 | 0.0115 |

**Table 1.** Segmentation accuracy (Dice coefficient, Jaccard index, Accuracy, Recall and Precision) on the Kvasir-SEG dataset

| CVC-ClinicDB | | | | | |
|---|---|---|---|---|---|
| **Method** | **DSC** | **Jaccard** | **Precision** | **Recall** | **Accuracy** |
| U-Net [7] (with our augmentations) | 0.7631 | 0.6169 | 0.7989 | 0.7303 | 0.9599 |
| HRNetV2 [14, 15] | 0.7776 | 0.6361 | 0.8260 | 0.7346 | 0.9629 |
| PraNet [32] (pre-trained) | 0.8742 | 0.7766 | 0.9608 | 0.8020 | 0.9780 |
| HarDNet-DFUS [17] (pre-trained) | 0.7279 | 0.5723 | 0.8945 | 0.6137 | 0.9586 |
| MSRF-Net [21] | 0.9060 | 0.8282 | 0.9547 | 0.8621 | 0.9842 |
| FCN-Transformer [19] (pre-trained) | 0.9327 | 0.8740 | 0.9728 | 0.8958 | 0.9886 |
| **OURS (no pre-training, 17 filters)** | 0.9450 | 0.8952 | **0.9488** | 0.9406 | 0.9903 |
| **OURS (no pre-training, 34 filters)** | **0.9478** | **0.9009** | 0.9468 | **0.9489** | **0.9907** |
| Standard Deviation (between methods) | 0.0837 | 0.1260 | 0.0622 | 0.1099 | 0.0131 |

**Table 2.** Segmentation accuracy (Dice coefficient, Jaccard index, Accuracy, Recall and Precision) on the CVC-ClinicDB dataset

| ETIS-LaribPolypDB | | | | | |
|---|---|---|---|---|---|
| Method | DSC | Jaccard | Precision | Recall | Accuracy |
| U-Net [7] (with our augmentations) | 0.7984 | 0.6969 | 0.8322 | 0.7724 | 0.9734 |
| HRNetV2 [14, 15] | 0.4720 | 0.3089 | 0.4645 | 0.4797 | 0.9433 |
| PraNet [32] (pre-trained) | 0.8827 | 0.7900 | 0.9825 | 0.8013 | 0.9877 |
| HarDNet-DFUS [17] (pre-trained) | 0.8662 | 0.7640 | 0.9708 | 0.7819 | 0.9869 |
| MSRF-Net [21] | 0.7791 | 0.6382 | 0.9191 | 0.6762 | 0.9797 |
| FCN-Transformer [19] (pre-trained) | 0.9163 | 0.8455 | 0.9633 | 0.8736 | 0.9915 |
| **OURS (no pre-training, 17 filters)** | 0.9324 | 0.8734 | **0.9539** | 0.9118 | 0.9930 |
| **OURS (no pre-training, 34 filters)** | **0.9354** | **0.8788** | 0.9309 | **0.9400** | **0.9931** |
| Standard Deviation (between methods) | 0.1433 | 0.1758 | 0.1620 | 0.1383 | 0.0156 |

**Table 3.** Segmentation accuracy (Dice coefficient, Jaccard index, Accuracy, Recall and Precision) on the ETIS-LaribPolypDB dataset

| CVC-ColonDB | | | | | |
|---|---|---|---|---|---|
| Method | DSC | Jaccard | Precision | Recall | Accuracy |
| U-Net [7] (with our augmentations) | 0.8032 | 0.7037 | 0.8100 | 0.8274 | 0.9807 |
| HRNetV2 [14, 15] | 0.6383 | 0.4687 | 0.5858 | 0.7010 | 0.9565 |
| PraNet [32] (pre-trained) | 0.9131 | 0.8401 | 0.9657 | 0.8659 | 0.9901 |
| HarDNet-DFUS [17] (pre-trained) | 0.7398 | 0.5870 | 0.9500 | 0.6057 | 0.9761 |
| MSRF-Net [21] | 0.8371 | 0.7198 | 0.8603 | 0.8151 | 0.9829 |
| FCN-Transformer [19] (pre-trained) | 0.9073 | 0.8304 | 0.9107 | 0.9040 | 0.9899 |
| **OURS (no pre-training, 17 filters)** | **0.9353** | **0.8785** | **0.9314** | **0.9392** | **0.9929** |
| **OURS (no pre-training, 34 filters)** | 0.9230 | 0.8571 | 0.9113 | 0.9351 | 0.9914 |
| Standard Deviation (between methods) | 0.0986 | 0.1367 | 0.1156 | 0.1099 | 0.0112 |

**Table 4.** Segmentation accuracy (Dice coefficient, Jaccard index, Accuracy, Recall and Precision) on the CVC-ColonDB dataset

| Trained on Kvasir-SEG tested on CVC-ClinicDB | | | | | |
|---|---|---|---|---|---|
| Method | DSC | Jaccard | Precision | Recall | Accuracy |
| U-Net [7] (with our augmentations) | 0.7010 | 0.5397 | 0.6640 | 0.7423 | 0.9441 |
| HRNetV2 [14, 15] | 0.7457 | 0.5945 | 0.7642 | 0.7280 | 0.9561 |
| PraNet [32] (pre-trained) | 0.7744 | 0.6319 | 0.9494 | 0.6539 | 0.9638 |
| HarDNet-DFUS [17] (pre-trained) | 0.5784 | 0.4068 | 0.5974 | 0.5605 | 0.9263 |
| MSRF-Net [21] | 0.6763 | 0.5109 | 0.6965 | 0.6572 | 0.9444 |
| FCN-Transformer [19] (pre-trained) | **0.8314** | **0.7114** | 0.8839 | **0.7848** | **0.9719** |
| **OURS (no pre-training, 17 filters)** | 0.8014 | 0.6686 | 0.8851 | 0.7321 | 0.9679 |
| **OURS (no pre-training, 34 filters)** | 0.8211 | 0.6965 | **0.8860** | 0.7650 | 0.9705 |
| Standard Deviation (between methods) | 0.0803 | 0.0975 | 0.1198 | 0.0692 | 0.0150 |

**Table 5.** Segmentation accuracy (Dice coefficient, Jaccard index, Accuracy, Recall and Precision) on the CVC-ClinicDB dataset, the models being trained on the Kvasir-SEG dataset

| Trained on CVC-ClinicDB tested on Kvasir-SEG | | | | | |
|---|---|---|---|---|---|
| Method | DSC | Jaccard | Precision | Recall | Accuracy |
| U-Net [7] (with our augmentations) | 0.5369 | 0.3670 | 0.4374 | 0.6951 | 0.8068 |
| HRNetV2 [14, 15] | 0.5531 | 0.3822 | 0.4242 | 0.7944 | 0.7931 |
| PraNet [32] (pre-trained) | 0.6852 | 0.5212 | 0.7647 | 0.6207 | 0.9130 |
| HarDNet-DFUS [17] (pre-trained) | 0.7272 | 0.5714 | 0.9180 | 0.6021 | 0.9261 |
| MSRF-Net [21] | 0.5152 | 0.3469 | 0.3939 | 0.7443 | 0.7742 |
| FCN-Transformer [19] (pre-trained) | **0.8800** | **0.7858** | **0.9659** | 0.8082 | **0.9645** |
| **OURS (no pre-training, 17 filters)** | 0.7525 | 0.6032 | 0.6873 | 0.8314 | 0.9119 |
| **OURS (no pre-training, 34 filters)** | 0.8251 | 0.7023 | 0.7740 | **0.8834** | 0.9396 |
| Standard Deviation (between methods) | 0.1285 | 0.1516 | 0.2121 | 0.0944 | 0.0698 |

**Table 6.** Segmentation accuracy (Dice coefficient, Jaccard index, Accuracy, Recall and Precision) on the Kvasir-SEG dataset, the models being trained on the CVC-ClinicDB dataset

## Ablation studies

In this part, the goal was to assess the efficiency of the proposed DUCK block compared to a standard convolutional block in a controlled, like-for-like test setup. Table 7 provides a comprehensive summary of the results derived from the ablation studies conducted.

| Tested on Kvasir-SEG | | | | | |
|---|---|---|---|---|---|
| **Block Variation** | **DSC** | **Jaccard** | **Precision** | **Recall** | **Accuracy** |
| DUCK-Net + Simple Convolution Blocks (17 filters) | 0.8722 | 0.7702 | 0.8656 | 0.8778 | 0.9593 |
| DUCK-Net + DUCK Blocks (17 filters) | 0.9343 | 0.8769 | 0.9350 | 0.9337 | 0.9789 |
| DUCK-Net + Simple Convolution Blocks (34 filters) | 0.8914 | 0.7921 | 0.8884 | 0.8845 | 0.9689 |
| DUCK-Net + DUCK Blocks (34 filters) | 0.9502 | 0.9051 | 0.9628 | 0.9379 | 0.9842 |

**Table 7.** Ablation studies results (Dice coefficient, Jaccard index, Accuracy, Recall and Precision) on the Kvasir-SEG dataset

The performance of the novel DUCK block was compared with a simple convolutional block in the context of the DUCK-Net architecture, using the Kvasir-SEG dataset. As shown in Table 7, the DUCK block consistently outperformed the simple convolutional block across all tested performance metrics. This advantage was evident in both the 17 and 34 filter size models. These findings indicate that the DUCK block significantly enhances DUCK-Net's performance, leading to more precise and accurate results. This analysis supports the utility of integrating the DUCK block within the DUCK-Net architecture for applications that demand high-performing convolutional blocks.

## Discussion

Supervised learning has proven to be effective for many tasks in the medical image domain, such as classification, detection, and segmentation. Advances in this field have been crucial for improving medical care, and developing high-performing models has played a central role in these advancements. Hence, developing methods that require minimal annotated data can be of great benefit to the clinical community.

This work presents a state-of-the-art (SOTA) model for automatic polyp segmentation in colonoscopy images. Through experiments, we demonstrate that our model outperforms existing models on various benchmarks, particularly in generalizability and handling polyps of varying shapes, sizes, and textures. This contributed to the automated processing of colonoscopy images can aid medical staff in lesion detection and classification. The model combines the strengths of wide information extraction from DeepLabV3+ [13] atrous convolutions with rich information extraction from a large yet efficient kernel in a separable module to localize the target polyp accurately.

Tables 1, 2, 3, and 4 show our experimental results on the SOTA polyp segmentation datasets Kvasir-SEG [28], CVC-ClinicDB [29], ETIS-LaribPolypDB [31], and CVC-ColonDB [30]. Our model outperforms all other architectures, highlighting its ability to learn key polyp features from small amounts of data. At the same time, Tables 5 and 6 show its capacity to generalize one dataset and apply it to a different one on Kvasir-SEG [28] and CVC-ClinicDB [29]. Even though it shows excellent results, it does not achieve SOTA results as FCN-Transformer [19] has the advantage of having extra training data (pre-training), which helps it generalize features in a less dataset-specific way. Furthermore, our model's capacity to handle real-world scenarios was demonstrated through the use of multiple datasets containing images that vary significantly from one another. These datasets include images from international patients with different backgrounds, representing a wide range of scenarios that our model could encounter in real-world applications.

Table 7 shows our results of the ablation studies conducted for the proposed DUCK block. Its consistent outperformance compared to a simple convolutional block supports the hypothesis that this novel structure enhances the effectiveness of image segmentation tasks. Future research might build on this foundation, exploring how the DUCK block performs in other architectures and tasks to further validate and leverage its advantages.

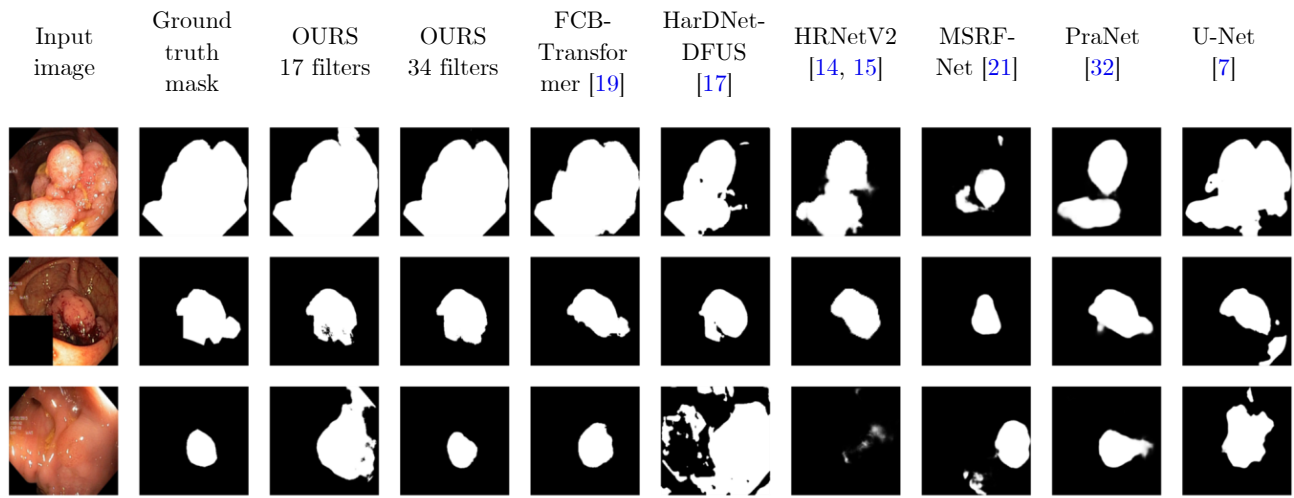

**Figure 7.** Comparison of predicted polyp masks

In Figure 7 we show three examples of polyp images from the Kvasir-SEG [28] test set, and we compare the predictions of our novel architecture "DUCK-Net", which was evaluated across different model sizes (17 and 34 filter size), to other existing architectures: FCB-Transformer [19], HarDNet-DFUS [17], HRNetV2 [14, 15], MSRF-Net [21], PraNet [32], and U-Net [7].

Regarding the computational complexity implications of integrating additional convolutional blocks within the DUCK block structure, we have to consider two main factors: computational cost and memory usage. Each new block means the network has to do more operations. For instance, a conventional convolutional layer with a kernel size of NxN has a computational complexity of $O(N^2)$. Furthermore, the residual blocks in DUCK would require the network to perform addition operations and potentially more non-linear operations such as sigmoid activation. On the other hand, memory usage also grows with the addition of each convolutional block. Every layer within a deep learning model must store its weights, gradients, and neuron activations, meaning that as more blocks are added, the model requires more memory to store these quantities during training and inference. These complexity considerations are why optimizations like dilated convolutions and separable convolutions are used, as they can provide similar representational power to standard convolutions but with fewer parameters and thus less computational cost. Ultimately, while using more blocks in DUCK will lead to more computational complexity, the advantage is that it allows the network to capture features at different scales and compensate for the drawbacks of different types of convolutions, which could improve the model's performance on complex tasks. Nevertheless, these benefits must be balanced against the increased resource requirements, particularly when deploying the model in resource-constrained environments.

While the model generally exhibits a high level of prediction accuracy, we have observed some limitations in its performance when dealing with polyps whose colors blend in with the background, resulting in indistinct borders. Further investigation is needed to address this issue and enhance the model's ability to locate and predict the borders of such polyps accurately.

## Conclusion

Based on the results presented in this paper, the DUCK-Net supervised convolutional neural network architecture can achieve state-of-the-art performance in polyp segmentation tasks in colonoscopy images. The model's encoder-decoder structure with a residual downsampling mechanism and custom convolutional block allows it to capture and process image information at multiple resolutions effectively. At the same time, data augmentation techniques help improve its overall performance. The DUCK-Net model demonstrates strong generalization capabilities and can achieve excellent results even with limited training data. Overall, the DUCK-Net architecture shows great potential for use in various segmentation tasks and warrants further investigation.

## Data availability

The randomly split datasets we tested the models on are publicly available at:

- **Datasets:** https://drive.google.com/drive/folders/1kg9XImzrd9PpTtleQSz6l8uq82LV1sjV?usp=share_link

The datasets used in this study are publicly available at:

- **Kvasir-SEG:** https://datasets.simula.no/kvasir-seg/
- **CVC-ClinicDB:** https://polyp.grand-challenge.org/CVCClinicDB/
- **ETIS-LaribpolypDB:** https://drive.google.com/drive/folders/10QXjxBJqCf7PAXqbDvoceWmZ-qF07tFi?usp=share_link
- **CVC-ColonDB:** https://drive.google.com/drive/folders/1-gZUo1dgsdcWxSdXV9OAPmtGEbwZMfDY?usp=share_link

## References

1. Siegel, R. L., Miller, K. D., Fuchs, H. E., & Jemal, A. Cancer statistics, 2022. *CA: A Cancer Journal for Clinicians.* https://doi.org/10.3322/caac.21708 (2022).
2. *American Cancer Society*. Colorectal cancer early detection, diagnosis, and staging. Retrieved from https://www.cancer.org/cancer/colon-rectal-cancer/detection-diagnosis-staging/detection.html (2021).
3. Shaukat, A. et al. ACG Clinical Guidelines: Colorectal Cancer Screening 2021. *The American Journal of Gastroenterology* **116(3),** 458-479. https://doi.org/10.14309/ajg.0000000000001122 (2021).
4. Pacal, I., Karaboga, D., Basturk, A., Akay, B., & Nalbantoglu, U. A comprehensive review of deep learning in colon cancer. *Computers in Biology and Medicine*, **126**, 104003. https://doi.org/10.1016/j.compbiomed.2020.104003 (2020).
5. Tharwat, M., Sakr, N. A., El-Sappagh, S., Soliman, H., Kwak, K., & Elmogy, M. Colon Cancer Diagnosis Based on Machine Learning and Deep Learning: Modalities and Analysis Techniques. *Sensors*, **22(23)**, 9250. https://doi.org/10.3390/s22239250 (2022).

6. Brigato, L., & Iocchi, L. A Close Look at Deep Learning with Small Data. *25th International Conference on Pattern Recognition (ICPR)* (pp. 2490-2497). https://doi.org/10.1109/ICPR48806.2021.9412492 (2021).
7. Ronneberger, O., Fischer, P., & Brox, T. U-Net: Convolutional Networks for Biomedical Image Segmentation. *Medical Image Computing and Computer-Assisted Intervention - MICCAI* (pp. 234-241). Springer International Publishing https://doi.org/10.1007/978-3-319-24574-4_28. (2015)
8. Long, J., Shelhamer, E., & Darrell, T. Fully convolutional networks for semantic segmentation. *IEEE Conference on Computer Vision and Pattern Recognition (CVPR)* (pp. 3431-3440). https://doi.org/10.1109/CVPR.2015.7298965 (2015).
9. Zhou, Z., Rahman Siddiquee, M. M., Tajbakhsh, N., & Liang, J. UNet++: A Nested U-Net Architecture for Medical Image Segmentation. *Deep Learning in Medical Image Analysis and Multimodal Learning for Clinical Decision Support* (pp. 3-11). Springer International Publishing. https://doi.org/10.1007/978-3-030-00889-5_1 (2018).
10. Jha, D. et al. ResUNet++: An Advanced Architecture for Medical Image Segmentation. I*EEE International Symposium on Multimedia (ISM)* (pp. 225-2255). https://doi.org/10.1109/ISM46123.2019.00049 (2019).
11. Li, X. et al. H-DenseUNet: Hybrid Densely Connected UNet for Liver and Tumor Segmentation From CT Volumes. *IEEE Transactions on Medical Imaging*, **37(12)**, 2663-2674. https://doi.org/10.1109/TMI.2018.2845918 (2018).
12. Chen, L.-C., Papandreou, G., Schroff, F., & Adam, H. Rethinking Atrous Convolution for Semantic Image Segmentation. CoRR, abs/1706.05587. Preprint at https://arxiv.org/abs/1706.05587 (2017).
13. Chen, L.-C., Zhu, Y., Papandreou, G., Schroff, F., & Adam, H. Encoder-Decoder with Atrous Separable Convolution for Semantic Image Segmentation. *Proceedings of the European Conference on Computer Vision (ECCV)*, 801-818 (2018).
14. Sun, K., Xiao, B., Liu, D., & Wang, J. *Proceedings of the IEEE/CVF Conference on Computer Vision and Pattern Recognition (CVPR)*, 5693-5703 (2019).
15. Sun, K., et al. High-Resolution Representations for Labeling Pixels and Regions. CoRR, abs/1904.04514. Preprint at http://arxiv.org/abs/1904.04514 (2019).
16. Diakogiannis, F. I., Waldner, F., Caccetta, P., & Wu, C. ResUNet-a: A deep learning framework for semantic segmentation of remotely sensed data. *ISPRS Journal of Photogrammetry and Remote Sensing*, **162**, 94-114. https://doi.org/10.1016/j.isprsjprs.2020.01.013 (2020).
17. Liao, T. Y., et al. HarDNet-DFUS: An Enhanced Harmonically-Connected Network for Diabetic Foot Ulcer Image Segmentation and Colonoscopy Polyp Segmentation. Preprint at https://arxiv.org/abs/2209.07313 (2022).

18. Duc, N. T., Oanh, N. T., Thuy, N. T., Triet, T. M., & Dinh, V. S. ColonFormer: An Efficient Transformer Based Method for Colon Polyp Segmentation. *IEEE Access*, **10**, 80575-80586. https://doi.org/10.1109/ACCESS.2022.3195241 (2022).
19. Sanderson, E., & Matuszewski, B. J. FCN-Transformer Feature Fusion for Polyp Segmentation. *Medical Image Understanding and Analysis* (pp. 892-907). Springer International Publishing, https://doi.org/10.1007/978-3-031-12053-4_65 (2022).
20. Wang, J., et al. Stepwise Feature Fusion: Local Guides Global. Preprint at https://arxiv.org/abs/2203.03635 (2022).
21. Srivastava, A., et al. MSRF-Net: A Multi-Scale Residual Fusion Network for Biomedical Image Segmentation. *IEEE Journal of Biomedical and Health Informatics*, **26(5)**, 2252-2263. https://doi.org/10.1109/JBHI.2021.3138024 (2022).
22. Deng, J., et al. ImageNet: A Large-Scale Hierarchical Image Database. *IEEE Conference on Computer Vision and Pattern Recognition (CVPR 2009)* (pp. 248-255). https://doi.org/10.1109/CVPR.2009.5206848 (2009).
23. Parmar, G., Zhang, R., & Zhu, J.-Y. On Aliased Resizing and Surprising Subtleties in GAN Evaluation. *Proceedings of the IEEE/CVF Conference on Computer Vision and Pattern Recognition (CVPR)* (pp. 11400-11410). https://doi.org/10.1109/CVPR52688.2022.01112 (2022).
24. Tieleman, T. & Hinton, G. Lecture 6.5-rmsprop: Divide the Gradient by a Running Average of Its Recent Magnitude. *COURSERA: Neural Networks for Machine Learning*, **4**, 26-31 (2012).
25. Abadi, M., et al. TensorFlow: Large-Scale Machine Learning on Heterogeneous Systems. Software available from https://www.tensorflow.org (2015).
26. Duchon, C. E. Lanczos Filtering in One and Two Dimensions. *Journal of Applied Meteorology and Climatology*, **18(8)**, 1016-1022. doi: https://doi.org/10.1175/1520-0450(1979)018%3C1016:LFIOAT%3E2.0.CO;2 (1979).
27. Buslaev, A., et al. Albumentations: Fast and Flexible Image Augmentations. *Information*, **11(2)**, 125. doi: https://doi.org/10.3390/info11020125">10.3390/info11020125 (2020).
28. Jha, D. et al. Kvasir-SEG: A Segmented Polyp Dataset. *MultiMedia Modeling. MMM 2020. Lecture Notes in Computer Science*, **11962**. Springer, Cham. https://doi.org/10.1007/978-3-030-37734-2_37 (2020).
29. Bernal, J., et al. WM-DOVA maps for accurate polyp highlighting in colonoscopy: Validation vs. saliency maps from physicians. *Computerized Medical Imaging and Graphics*, **43**, 99-111. doi: https://doi.org/10.1016/j.compmedimag.2015.02.007 (2015).

30. Vázquez, D., et al. A Benchmark for Endoluminal Scene Segmentation of Colonoscopy Images. *Journal of Healthcare Engineering.* doi: https://doi.org/10.1155/2017/4037190 (2017).
31. J. Bernal, et al. Comparative Validation of Polyp Detection Methods in Video Colonoscopy: Results From the MICCAI 2015 Endoscopic Vision Challenge. *IEEE Transactions on Medical Imaging*, **36**, 6, pp. 1231-1249. doi: https://doi.org/10.1109/TMI.2017.2664042 (2017).
32. Fan, DP. et al. PraNet: Parallel Reverse Attention Network for Polyp Segmentation. *Medical Image Computing and Computer Assisted Intervention. Lecture Notes in Computer Science*, **12266**. Springer, Cham. https://doi.org/10.1007/978-3-030-59725-2_26 (2020).
33. Vaswani, A., et al. Attention is All You Need. *Advances in Neural Information Processing Systems.* Curran Associates, Inc. Retrieved from https://proceedings.neurips.cc/paper/2017/file/3f5ee243547dee91fbd053c1c4a845aa-Paper.pdf (2017).
34. Dosovitskiy, A., et al. An Image is Worth 16x16 Words: Transformers for Image Recognition at Scale. Preprint at https://arxiv.org/abs/2010.11929 (2020).
35. Tan, M., & Le, Q. EfficientNetV2: Smaller Models and Faster Training. *Proceedings of the 38th International Conference on Machine Learning* (pp. 10096-10106). Retrieved from http://proceedings.mlr.press/v139/tan21a/tan21a.pdf (2021).
36. Krizhevsky, A. & Hinton, G. Learning multiple layers of features from tiny images. *Technical Report, 2009.* Retrieved from https://www.cs.toronto.edu/~kriz/learning-features-2009-TR.pdf (2009).
37. Chaurasia, A., & Culurciello, E. LinkNet: Exploiting encoder representations for efficient semantic segmentation. *IEEE Visual Communications and Image Processing (VCIP)* (pp. 1-4). doi: https://doi.org/10.1109/VCIP.2017.8305148 (2017).
38. Alom, M. Z., et al. A State-of-the-Art Survey on Deep Learning Theory and Architectures. *Electronics*, **8(3)**, 292. doi: https://doi.org/10.3390/electronics8030292 (2019).
39. Chen, X., & Lin, X. Big Data Deep Learning: Challenges and Perspectives. *IEEE Access*, **2**, 514-525. doi: https://doi.org/10.1109/ACCESS.2014.2325029 (2014).
40. Sun, H., Xu, C., & Suominen, H. Analyzing the Granularity and Cost of Annotation in Clinical Sequence Labeling. CoRR, abs/2108.09913. Preprint at https://arxiv.org/abs/2108.09913 (2021).
41. H. Seo, et al. Modified U-Net (mU-Net) With Incorporation of Object-Dependent High Level Features for Improved Liver and Liver-Tumor Segmentation in CT Images. *IEEE Transactions on Medical Imaging*, **39**, 5, pp. 1316-1325. doi: https://doi.org/10.1109/TMI.2019.2948320 (2020).

## Acknowledgements

The development and initial experiments were run on a server provided by CLOUD-UT from the Technical University of Cluj-Napoca. The final results were run on a more powerful server provided by Dr. Bogdan Mihai Neamtu.

## Author contributions

R.D. developed the model architecture, the custom "DUCK" block and contributed to data augmentation. D.P. performed the data preprocessing, data augmentation, data analysis and training/testing the model. C.C. contributed with theoretical information about existing ML models and implemented, tested and compared different architecture types. All authors contributed with ideas during development, discussed the results and provided help writing the manuscript.

## Competing interests

The authors declare no competing interests.
24